\documentclass[conference]{IEEEtran}
\usepackage{times}

\usepackage[sort, numbers]{natbib}

\usepackage{multicol}
\usepackage[bookmarks=true]{hyperref}
\usepackage[nolist]{acronym}
\usepackage{mathtools}
\DeclarePairedDelimiter\set\{\}
\usepackage{amsfonts}
\usepackage{siunitx}
\usepackage{todonotes}
\usepackage[bookmarks=true]{hyperref}
\usepackage{balance}
\usepackage{hyperref}
\usepackage{subfigure}

\hypersetup{
    colorlinks=true,
    linkcolor=blue,
    filecolor=magenta,      
    urlcolor=cyan,
    hyperindex=True,
 urlcolor=cyan}
 
\newlength{\defaulttextfloatsep}
\newlength{\defaultintextsep}
\begin{document}

\title{Vision-Based Goal-Conditioned Policies for Underwater Navigation in the Presence of Obstacles}




%
\author{\authorblockN{Travis Manderson\authorrefmark{1},
Juan Camilo Gamboa\authorrefmark{1},
Stefan Wapnick\authorrefmark{1}, 
Jean-Fran\c{c}ois Tremblay\authorrefmark{1},\\
Florian Shkurti\authorrefmark{2},
Dave Meger\authorrefmark{1} and
Gregory Dudek\authorrefmark{1}}
\authorblockA{\authorrefmark{1}Mobile Robotics Laboratory, 
School of Computer Science,
McGill University, Montreal, Canada\\
\authorblockA{\authorrefmark{2}Robot Vision \& Learning Lab, Department of Computer Science, University of Toronto, Canada}
Email:  {\tt\small \{travism,gamboa,swapnick,jft,dmeger,dudek\}@cim.mcgill.ca, florian@cs.toronto.edu}}
}


\maketitle

\begin{abstract}
We present Nav2Goal, a  data-efficient and end-to-end learning method for  goal-conditioned visual navigation. Our technique is used to train a navigation policy that enables a robot to navigate close to sparse geographic waypoints provided by a user without any prior map, all while avoiding obstacles and choosing paths that cover user-informed regions of interest. Our approach is based on recent advances in conditional imitation learning. General-purpose safe and informative actions are demonstrated by a human expert. The learned policy is subsequently extended to be goal-conditioned by training with hindsight relabelling, guided by the robot's relative localization system, which requires no additional manual annotation. We deployed our method on an underwater vehicle in the open ocean to collect scientifically relevant data of coral reefs, which allowed our robot to operate safely and autonomously, even at very close proximity to the coral. Our field deployments have demonstrated over a kilometer of autonomous visual navigation, where the robot reaches on the order of 40 waypoints, while collecting scientifically relevant data. This is done while travelling within 0.5 m altitude from sensitive corals and exhibiting significant learned agility to overcome turbulent ocean conditions and to actively avoid collisions. 
\end{abstract}

\IEEEpeerreviewmaketitle

\section{Introduction}
\label{sec:intro}

This paper describes {\em Nav2Goal}\footnote{Project details can be found at \url{http://www.cim.mcgill.ca/mrl/nav2goal/}}, a robust visual navigation system, trained through goal-conditioned imitation learning, that learns safe, reactive behaviours for close-range robotic inspection of challenging geographic features using a relatively small annotation budget. Our methodology is generic enough to allow easy synthesis of goal-directed navigation, informative
path planning (i.e. relevant data collection), and collision avoidance. This navigation system could, for example, be used by biologists to automate data collection in unstructured environments such as coral reefs, which is a critical need for environmental monitoring and understanding.

Our approach begins by training, through behavioural cloning, a safe policy capable of seeking scientifically desirable observations. This approach requires an expert to label steering commands for our vehicle, assigning a pitch and yaw angle to each image in a training dataset extracted from previous exploration runs with the vehicle. Using this dataset to train a policy, the robot is able to navigate safely and prefers observations of coral, but it lacks the ability to reach goals (waypoints).

Our proposed method addresses this shortcoming by augmenting this learned policy with goal-seeking behaviour using hindsight experience relabelling~\cite{NIPS2019_9667}, 
adapted to our field robotics setting. To this end, we collected a dataset of trajectories from executions of our safe and coral-preferring policy to enable goal-seeking navigation behaviours within a large radius relative to the robot. We use a vision-based state estimator to track the robot's position. In this way, each sub-segment of the recorded trajectories is automatically labelled for goal-directed navigation from the beginning of the sub-segment to the end. By re-training a conditional imitation learning policy on this dataset, we achieved safe, coral-seeking and goal-directed navigation behaviour without any additional annotation effort. 
\begin{figure}[t]
  \centering
  \includegraphics[width=1.0\linewidth]{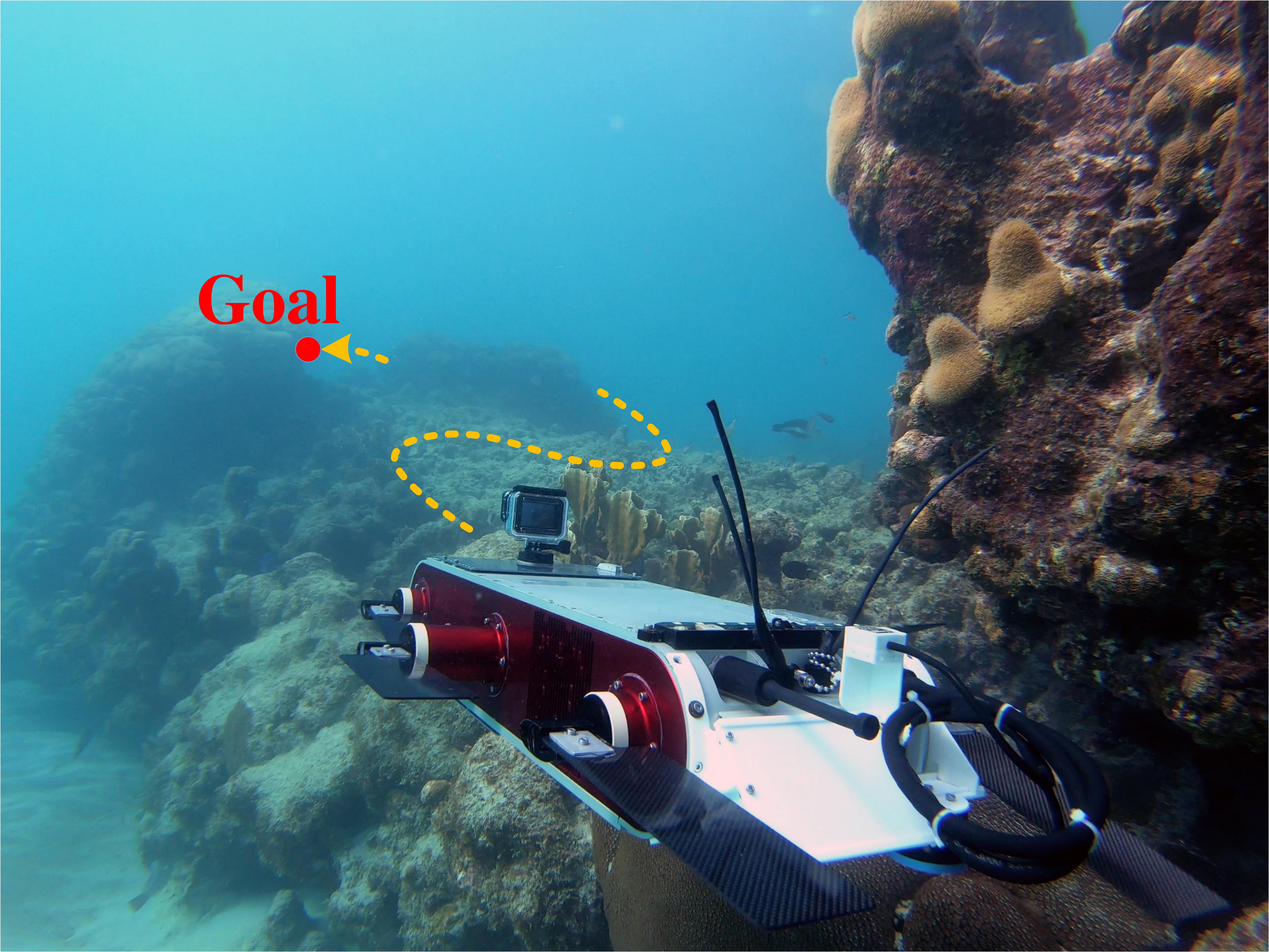}
  \caption{Robotic vehicle swimming autonomously towards a goal using vision, while simultaneously following the user-specified objective of visiting coral regions along the way.}
\label{fig:robot_into}
\end{figure}
The resulting goal-conditioned policy from our method can be used to navigate to an arbitrary set of waypoints (relative to the robot's frame of reference) provided by the user.

\begin{figure*}[t!]
  \centering
  \includegraphics[width=0.95\linewidth]{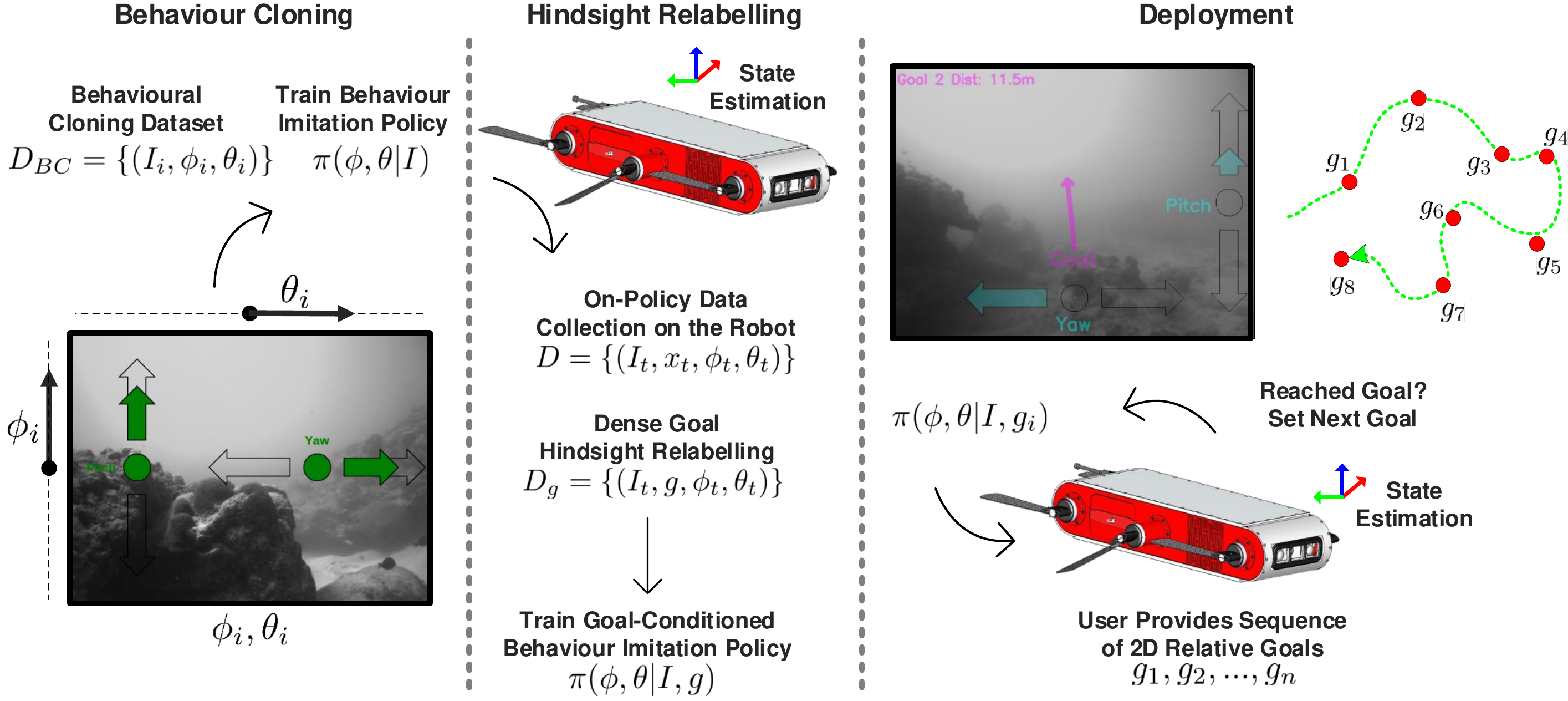}
  \caption{System overview consisting of 3 major parts. In the first phase, a non-goal conditional policy is learned through behavioural cloning, using expert labeled images for obstacle avoidance and scientific data collection. Second, we run the policy learned with some exploration bias and generate dense, goal-conditioned action labels, from which we can train a goal-conditioned policy using hindsight relabelling. Third, in the deployment phase, the robot is given a sequence of waypoints to navigate to while simultaneously satisfying higher-level tasks, such as surveying coral. }
\label{fig:block_diagram}
\end{figure*}

We demonstrate our approach on an \ac{AUV} both in simulation and in a field robotics setting in the open ocean. Goal-conditioned data is generated by executing a behaviour policy trained to avoid obstacles while staying above coral reef. In Fig.~\ref{fig:robot_into}, we show a representative example of a path segment that our approach followed by successfully reaching waypoints, while simultaneously preferring to navigate over a coral reef to capture relevant data.

In summary, the main contributions of our paper are:
\begin{itemize}

    \item A vision-based method to robustly perform low-level behaviours, such as obstacle avoidance in underwater environments, with a relatively small amount of training data from an expert. 
    
    \item A method that adaptively combines low-level behaviours, such as obstacle avoidance, with high-level behaviours, such as goal-directed behaviour and exploratory scientifically-relevant observation.
    
    \item The first end-to-end visual navigation system that shows successful goal-conditioned imitation learning behaviour deployed in a marine environment. 
    Our robot autonomously swam over a kilometer collision-free while reaching on the order of 40 waypoints, while observing coral and without relying on a global map.
\end{itemize}

\section{Background and related work}
\label{sec:related}
Our work is related to existing literature on vision-based imitation learning for autonomous navigation in the field.

\textbf{Imitation Learning:} Although simple imitation can be done via supervised learning (behavioural cloning)~\cite{pomerleau1989alvinn} on observation and action pairs from an expert, it is not robust to distribution shift. 
The learned policy might take the system to regions of the state space that have not been seen in the training set, 
leading to poor generalization. In fact, the accumulated error in behavioural cloning exhibits quadratic dependence with 
respect to the length of the sequence of actions executed. Behavioural cloning can be made more robust in practice by 
augmenting the set of observations for which the policy is valid \cite{bojarski_nvidia_driving, Giusti2016, dart_imitation_learning}
or by trying to match the distribution of trajectories that the expert demonstrated~\cite{gail}. However, the underlying problem of 
compounding errors during distribution shift remains. DAgger~\cite{ross2011reduction} addresses this problem by 
iteratively querying the expert on states that are visited by the learned policy during evaluation. Aside from the accumulated 
error scaling linearly with respect to the time horizon, this technique has been demonstrated on the task of visual navigation 
on real robot systems~\cite{ross_monocular_uav}. More recent methods have combined human demonstrations and imitation learning~\cite{singh2019} with end-to-end reinforcement learning on visual data \cite{end_to_end} as another way to handle distribution shift.  

\textbf{Learning for Goal-Conditioned Navigation:} While typical imitation learning scenarios consider a single task (for instance, lane following or navigation to a fixed goal)~\cite{Giusti2016, Loquercio2018, Smolyanskiy2017}, we also want to be able to learn goal-conditioned policies. Conditional imitation learning has seen significant research activity in the last few years~\cite{conditional_imitation_learning_codevilla, NIPS2019_9667,rhinehart2018deep, held_changing_reward_envs, end_to_end_motion_planning_ethz} and has many connections to goal-conditioned reinforcement learning \cite{universal_value_functions, Kaelbling93b, HER}, particularly in the batch case. Imitation approaches that generally do not rely on end-to-end learning but can handle multiple goals include \cite{barfoot_reusable_paths}.

\textbf{Underwater Navigation:} Most existing underwater navigation systems rely on a combination of acoustic, magnetic, and inertial sensors \cite{underwater_stefan_williams, terrain_aided_nav}. This technique enables localization over large distances in terms of space and time (potentially over the course of months). These systems are tailored to deep bathymetry and typically avoid navigation very near the seafloor. Vision-based underwater navigation systems \cite{manderson2018vision, eustice_pizarro_singh_1, singh_roman_pizarro_eustice, Singh2004, yogi_ijrr} tailored to the close-range exploration of the seafloor provide a rich, high-resolution source of observations for marine ecosystems. 
Some recent work has reported trajectory
optimization in 3D, but unlike this work
needs a prior map and has
only been validated in practice in
controlled man-made environments~\cite{XanthidisICRA2020}.

\textbf{Informative Path Planning:} Our work is close in both spirit and motivation to information-gathering behaviours in path planning \cite{sacbp, informative_planning_binney, info_gathering_arora, nonmyopic_informative_path_planning, liam_adaptive_path_planning, Charrow2015InformationTheoreticPW}. Most existing methods in this category need to estimate the surrounding map while executing variants of frontier-based exploration. In contrast, our work does not assume a map and does not need to perform exploration-exploitation within that map, which is sometimes called Simultaneous Planning, Localization and Mapping (SPLAM) \cite{labbe2018long}. This is because we directly make use of reactive policies, as opposed to having a pipeline that includes perception, mapping, path planning and tracking.

\section{Approach}
\label{sec:approach}
Our method called Nav2Goal is designed with the purpose of extracting a goal-conditioned policy, useful for navigation, from a lower level behaviour policy trained via imitation learning. 
The purpose of this design is such that high-level behaviours (for instance, relevant scientific data collection via waypoint seeking) are harmoniously built on top of low-level behaviours (e.g. obstacle avoidance reactive target following). An overview of the method is illustrated in Fig. \ref{fig:block_diagram}. We provide a summary of the three phases involved below: 

First, in the behavioural cloning phase, a user labels image frames (from video collected by the robot) with actions (corresponding to desired pitch and yaw changes) 
for the purposes of obstacle avoidance and scientific data collection. The objective of this labelling is to train a policy that enables the robot to avoid obstacles when close to them, and otherwise, to direct its camera towards parts of the scene that are relevant to the user's desires. 
Second, in the exploratory navigation and hindsight relabelling phase, we run the policy learned above with some exploration bias to collect trajectories demonstrating the desired behaviour. From these trajectories, relying on visual odometry, we automatically generate input-label pairs from which we can train a goal-conditioned policy.
Third, in the deployment phase, the robot is given a sequence of waypoints, and executes the goal-conditioned policy to approximately reach each waypoint, while satisfying the desired behaviour encoded in the original policy learned by imitation. An overview of these three phases is shown in Fig.~\ref{fig:block_diagram}.  

\subsection{Behavioural Cloning}
\label{sec:behavior_cloning}
Our method follows a similar approach as the behavioural cloning method of \cite{manderson2018vision}, with a slight variation in the labelling objective. As opposed to \cite{manderson2018vision}, more emphasis was placed on collision avoidance during the labelling process. Our dataset consists of images collected with the robot's camera, which are labelled with a desired change in heading, representing the action the user thinks that the robot should have taken. A \ac{CNN} model based on the ResNet-18 architecture~\cite{He_2016_CVPR} is trained on single image inputs to predict these relative heading changes in yaw and pitch. Such an architecture has been used previously for the task of visual navigation in forests trails~\cite{Smolyanskiy2017}. Yaw and pitch values are each encoded as a discrete set of seven class labels. Concrete Dropout~\cite{Gal2017ConcreteB} is used for regularization and uncertainty estimation.

We denote the yaw and pitch categorical distributions predicted by the policy network by $\mathbf{f}^{(\theta)}$ and $\mathbf{f}^{(\phi)}$ respectively. The seven class prediction labels are centered with mean of 0 and are described by the set $C=\lbrace-3,-2,-1,0,1,2, 3\rbrace$. They represent the desired, unscaled yaw or pitch change with respect to the current image frame, with negative classes corresponding to clockwise/downwards and positive classes corresponding to anti-clockwise/upwards heading changes. 
Seven classes were chosen as they were found to give sufficient fidelity in control actions while still constituting a small enough set for a machine learning model to learn easily. Furthermore, we performed \emph{label smoothing} on the ground-truth labels during training as is suggested in~\cite{Smolyanskiy2017}.
The following loss is used to train the policy:
\begin{align}
    \mathcal{L}\left(D,\mathbf{w}\right) &= \mathcal{L}_{\mathrm{pred}}\left(D,\mathbf{w}\right) + \lambda_1\mathcal{L}_{\mathrm{reg}}\left(D,\mathbf{w}\right)\\
    \mathcal{L}_{\mathrm{pred}} &= \sum_{(\mathbf{I}_i, \mathbf{\theta}_i, \mathbf{\phi}_i) \in D} l(\mathbf{f}^{(\theta)}_i, \mathbf{\theta}_i, \mathbf{w}) + l(\mathbf{f}^{(\phi)}_i, \mathbf{\phi}_i, \mathbf{w})
\end{align}

\noindent where $\mathbf{f_{i}}^{(\theta)}$ and $\mathbf{f_{i}}^{(\phi)}$ represent the network output heads for yaw and pitch action predictions of the policy $\pi(\theta,\phi|I)$ learned in the behavioural cloning phase, which is parameterized by weights $\mathbf{w}$. The dataset $D$ corresponds to $D_{BC}$, the expert-labeled images and actions, as shown in Fig.~\ref{fig:block_diagram}. The image input is denoted as ${\mathbf{I}_i}$ while ${\theta_{i}}$ and ${\phi_{i}}$ represent the one-hot encoding of the yaw and pitch ground-truth labels. The predictive loss for yaw $l(\mathbf{f}^{(\theta)}_i, \mathbf{\theta}_i, \mathbf{w})$ consists of the sum of a multi-class cross-entropy loss (between network predictions and ground-truth labels) and a regularization term that penalizes over confident predictions:
\begin{equation}
l(\mathbf{f}^{(\theta)}_i, \mathbf{\theta}_i, \mathbf{w}) = -\sum_j \theta_{ij} \log f^{(\theta)}_{ij} - \lambda_2 \sum_j f^{(\theta)}_{ij} \log f^{(\theta)}_{ij}
\end{equation}
\noindent where the $j^{\text{th}}$ index specifies the probability of each class. The loss for pitch predictions $l(\mathbf{f}^{(\phi)}_i, \mathbf{\phi}_i, \mathbf{w})$ is defined in the same way.
The $\mathcal{L}_{\mathrm{reg}}$ loss term corresponds to the KL-divergence term introduced in Concrete Dropout~\cite{Gal2017ConcreteB}.

During inference, the 7-class discrete network predictions for yaw and pitch are converted to low-level actuator commands executed by the robot's motion controllers. Temporal smoothing is applied on the network action predictions to reduce oscillatory motion caused by rapidly fluctuating motor commands. Dropout is left enabled during inference, as a way to introduce diversity in the actions taken by the robot for inputs that are out of the training distribution. 

\subsection{Goal Conditioned Model}
\label{sec:model}
 We are interested in ensuring that the strategies used by the robot to reach a goal are aligned with the behaviour policy. In our work, we achieve this by training a goal-conditioned policy $\pi(\theta,\phi|I,g)$ with automatic hindsight relabelling of experience gathered with the behaviour policy we wish to imitate, as shown in Fig.~\ref{fig:block_diagram}.
 
\begin{figure}[t]
  \centering
  \includegraphics[width=1.0\linewidth]{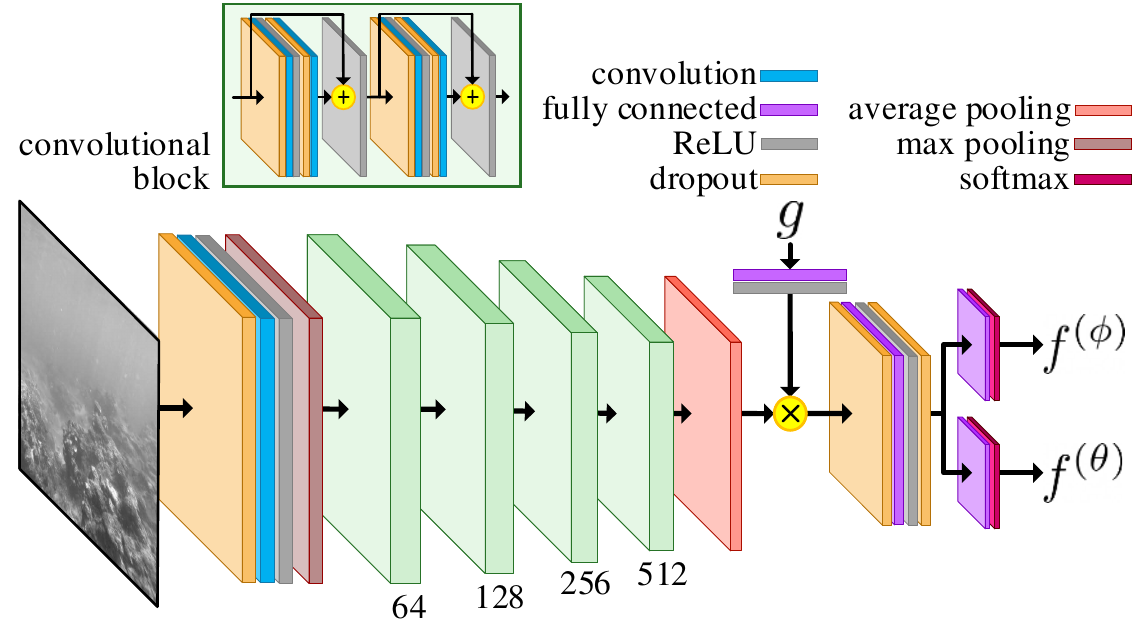}
  \caption{Goal-conditioned network that predicts actions for the current image frame to reach a goal while simultaneously following coral. The network is composed of a ResNet-18 \ac{CNN} which reads a single input image and its feature map output is combined with an input goal. Two softmax heads predict yaw and pitch categorical distributions, $\mathbf{f}^{(\theta)}$ and $\mathbf{f}^{(\phi)}$ respectively, over 7-class outputs each.}
\label{fig:gc_aqua_net}
\end{figure}

Our proposed goal-conditioned navigation policy extends the behavioural cloning architecture described in section~\ref{sec:behavior_cloning}. 
The previous policy is augmented with a two-dimensional vector input describing the goal as shown in Fig.~\ref{fig:gc_aqua_net}. Goals are described relative to the robot's current frame instead of absolute world coordinates. Since there is little variation in environment elevation, the goal is represented in planar coordinates for simplicity. Although the robot still navigates in 3-dimensions and must predict pitch values for proper collision avoidance. Goals described by both Cartesian and polar coordinates were explored during evaluation as a hyper-parameter. In our architecture, the goal input is fed through a dense network layer before being combined with the final flattened ResNet-18 ~\ac{CNN} feature map output. The goal input and image feature map are combined by either multiplication or concatenation. The choice of combination method was likewise tuned as another hyper-parameter during our experiments.

The network objective function remains unchanged from the original non-goal conditioned model described in~\ref{sec:behavior_cloning} with the exception that there is now an additional goal input.

\subsection{Hindsight Relabelling}
\label{sec:hindsight_relabelling}
During operation, state estimation running on-board the robot is used to obtain relative pose data (described in Section~\ref{sec:state_estimation}) and generate datasets of location-aware trajectories. Trajectory elements are described by the tuple $\langle\mathbf{I}_i, \mathbf{x}_i, \mathbf{\theta}_i, \mathbf{\phi}_i\rangle$, where $\mathbf{I}_i$ is the current forward-facing image, $\mathbf{x}_i$ is the estimated pose, and $\mathbf{\theta}_i, \mathbf{\phi}_i$ are the applied yaw and pitch actions respectively. These trajectories can be broken up into examples for the goal conditioned navigation task: \emph{which action should the robot execute to arrive at a particular relative location?} For this task, the inputs to the prediction model are the current image and goal relative to the robot's current location while the output is the action that the robot executed in that situation. 


During training, goals are sampled from the collected trajectories in a method similar to~\cite{HER}.
For each training mini-batch, a random set of trajectories and timesteps within those trajectories are selected from the dataset. Each sampled item is randomly associated with a future timestep which is interpreted as a goal. A maximum timestep difference $\tau$ between the two associated states is enforced to ensure that the goal is not excessively far away. Specifically, for a sampled trajectory timestep $t$ represented by the tuple $(\mathbf{I}_t, \mathbf{x}_t, \mathbf{\theta}_t, \mathbf{\phi}_t)$, a goal is associated as:
\begin{align}
\mathbf{g}_t &= \text{diff}(\mathbf{x}_t, \mathbf{x}_{t+\Delta{t}})\\
\Delta{t} &\sim \mathrm{Uniform\left[1, \min{(\tau, T-t)}\right]}
\end{align}
\noindent where $T$ is the trajectory length and $\tau$ is the maximum allowable timestep difference for sampling. After association with a goal from hindsight relabelling, training tuples in the format $\langle\mathbf{I}_i, \mathbf{g}_i, \mathbf{\theta}_i, \mathbf{\phi}_i\rangle$ are applied to the goal-conditioned network described in~\ref{sec:model}.

\begin{figure}[t!]
  \centering
  \includegraphics[width=1.0\linewidth]{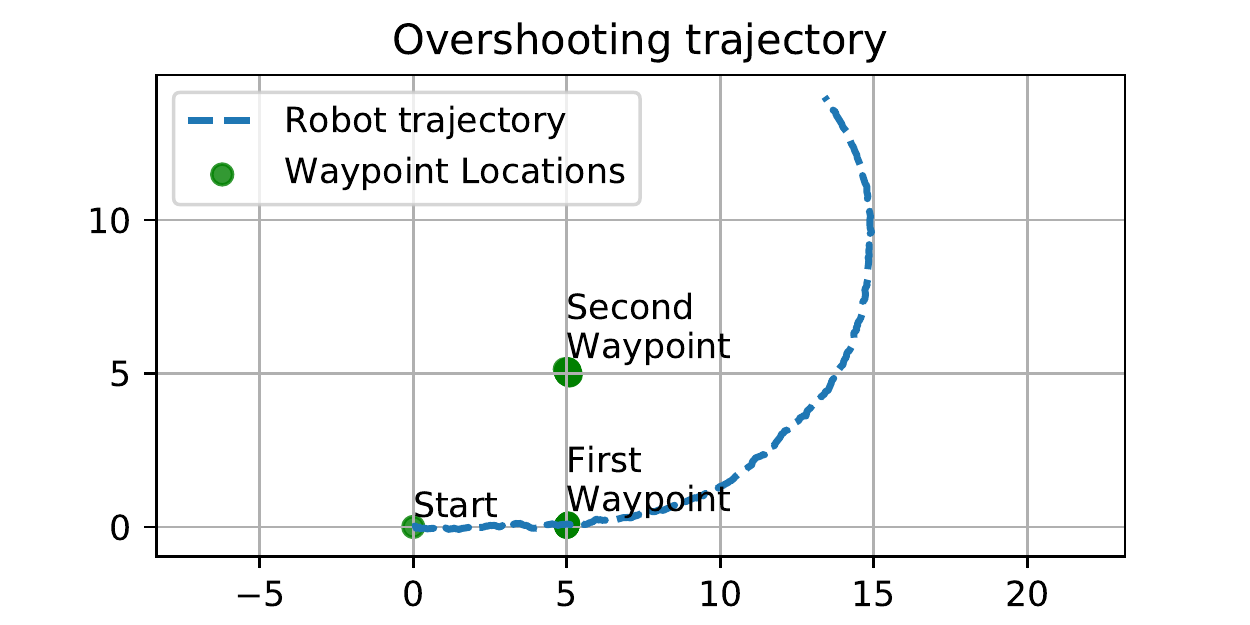}
  \caption{Overshooting trajectory due to the robot's constant forward speed used in these experiments. While the robot moves in the correct direction, it cannot turn fast enough.}
\label{fig:L_rect}
\end{figure}

\subsection{Waypoint Selection and Controller}
\label{sec:waypoint}

Once training has completed, our learned policy can be executed to guide the robot through a trajectory of relative position waypoints that define a mission. The waypoints are selected sequentially to become the active conditioning goal for policy execution. The robot's current image and this goal are sufficient inputs to compute the steering action to apply in real time. Once the robot approaches within a pre-defined threshold of the current waypoint, it is considered achieved, and the subsequent waypoint is set as the active goal. 

Given that the motion of the robot is constrained to flat
swimming\footnote{The robot tries to keep its roll angle constant at 0, and pitch targets are centered around 0 pitch.} and that the policy is reactive (i.e. it only considers the current image and goal for deciding whether to turn), not all possible motions are represented within our datasets. Furthermore, since the robot is swam at a constant speed, the turning radius is limited by the drag forces acting on the robot's body. As an example of the limitation of the waypoint controller, consider the situation depicted in Fig.~\ref{fig:L_rect}, where the robot was tasked with reaching a waypoint in front of it and then turning to the left 90 degrees for the next waypoint. While the robot takes the correct action (turning to the left), it is unable to turn fast enough to hit the next waypoint.

\begin{figure}[t]
  \centering
  \includegraphics[width=1.0\linewidth]{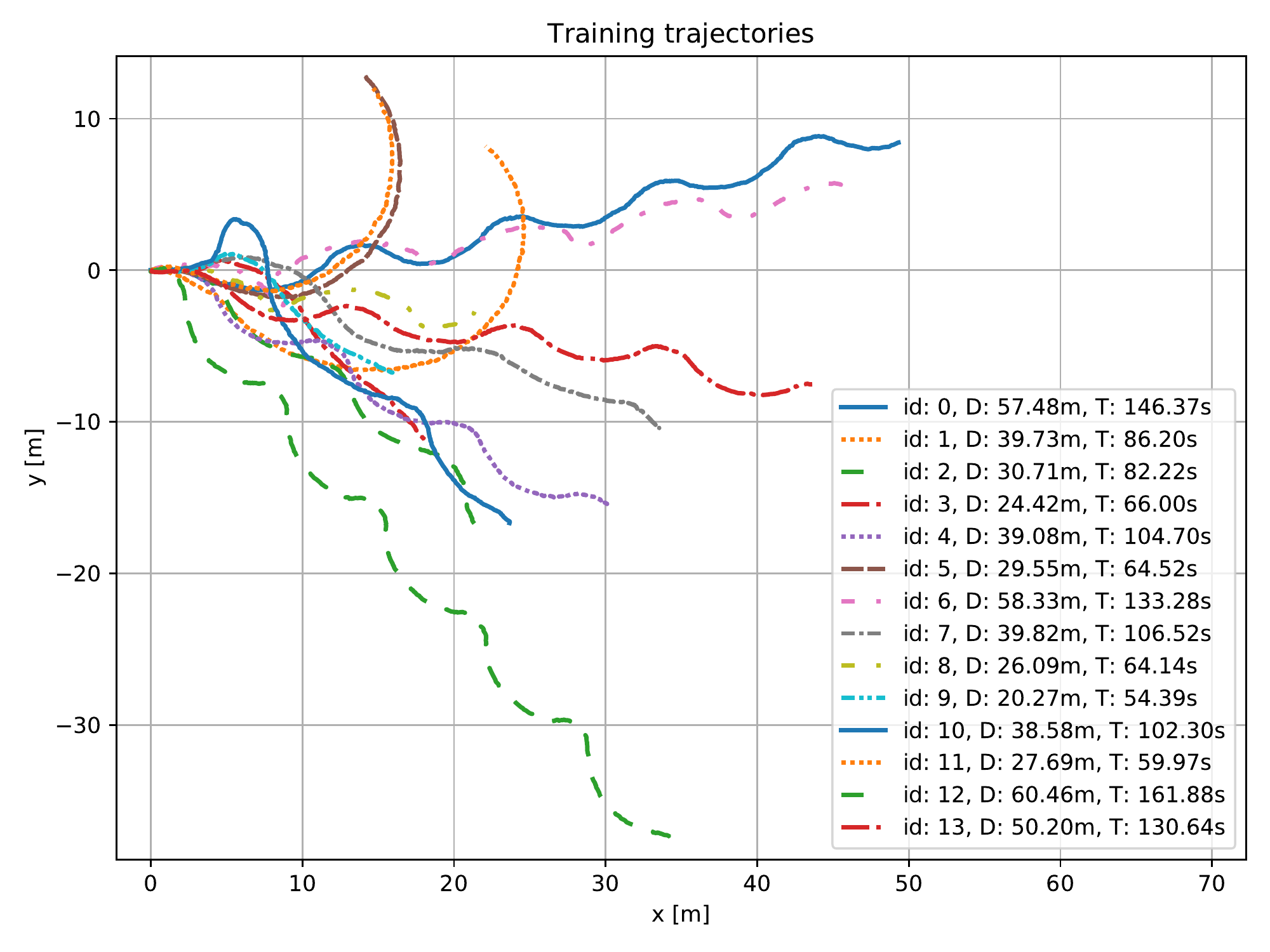} \\
  \includegraphics[width=1.0\linewidth]{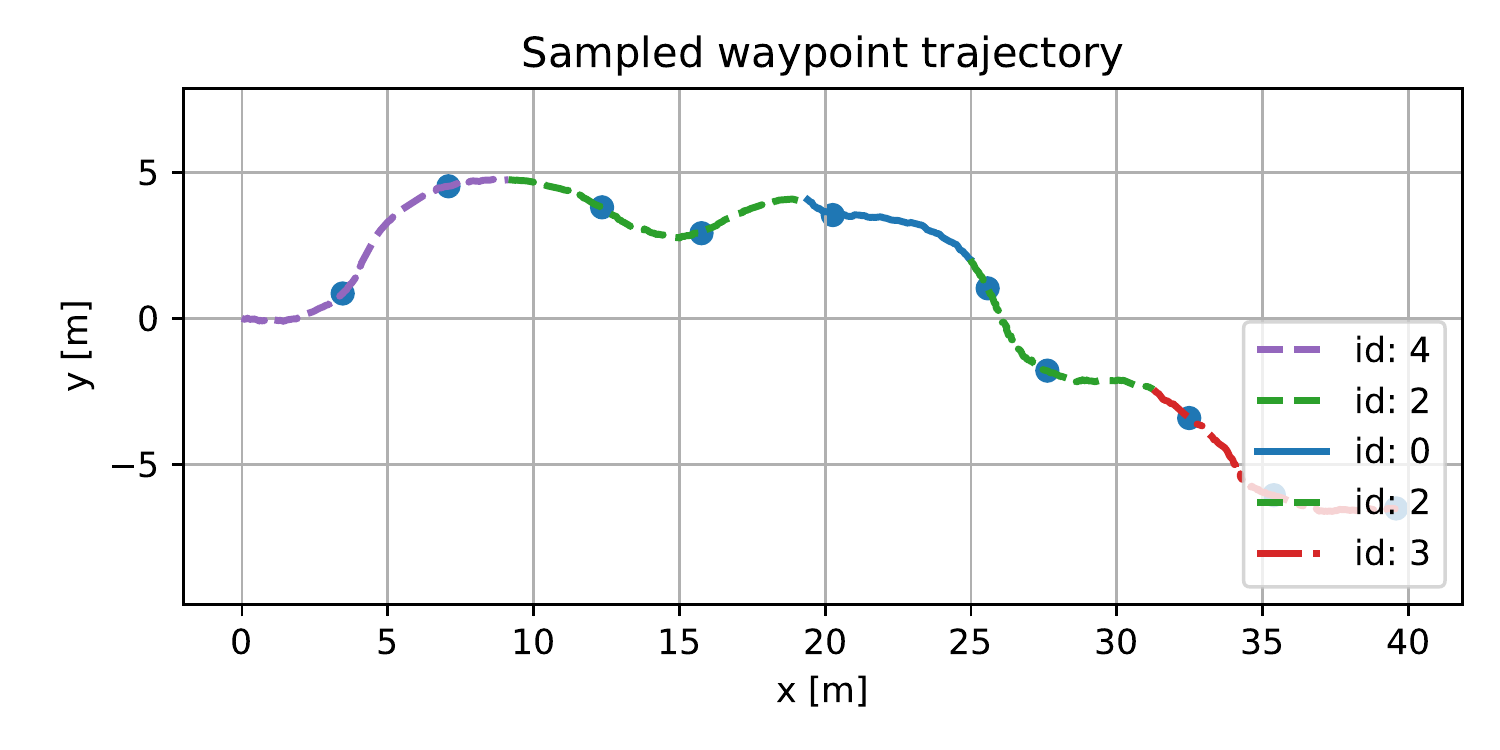}
  \caption{Top: Ensemble of training trajectories collected during deployment of the behaviour policy, described in Section~\ref{sec:results}. Bottom: Example trajectory created with our sampling procedure, which ensures that the trajectory is realizable by the robot. }
\label{fig:expl_traj}
\end{figure}

To deal with this situation, we construct waypoint trajectories by splicing together trajectory segments that exist in our dataset. This is done to ensure that the waypoint trajectories are executable by the robot, since they were actually executed during data collection with the behaviour policy. Fig.~\ref{fig:expl_traj} shows an example of one such trajectory and the corresponding dataset that was used to generate it.

While in this work we use this algorithm for generating fixed samples of waypoint trajectories, we can use a similar scheme with a sampling-based planner (e.g. RRTs~\cite{rrt2001}) where we ensure that trajectory segments are taken from the trajectory distribution explored by the behaviour policy.

%
%
%
%

\section{Experimental Methodology}

In order to validate the effectiveness of our approach in real-world conditions, we have targeted a sample task with relevance in modern ocean science. Specifically, we consider a small \ac{AUV} navigating in a region containing mixed coral, such that arriving at waypoints along coral-rich paths while avoiding large regions of sand is of scientific utility. This section describes the robot platform and supporting systems components utilized for conducting these trials.

\subsection{Underwater Autonomous Vehicle}
\label{sec:aqua}
\begin{figure}[t!]
  \centering
  \includegraphics[width=1.0\linewidth]{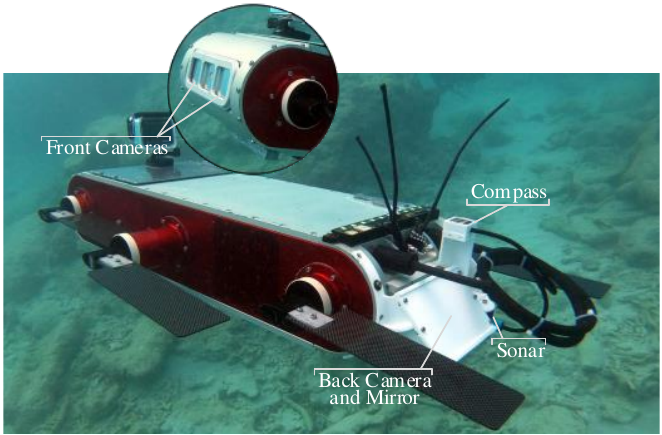}
  \caption{Underwater robot platform including 2 front cameras, a downward-facing rear camera, sonar, and a compass.}
\label{fig:robot_platform}
\end{figure}

Our experimental robot is a human-portable swimming hexapod~\cite{Rekleitis2005d}
weighing 15 kg and measuring 60 cm in length with six flippers for propulsion. Internally the robot contains three computers: 1) an embedded computer running a real-time operating system for low-level control such as driving the motors and reading encoders and other health-related sensors, 2) a small Intel i3 based computer for running high-level computation tasks such as image processing and navigation algorithms, and 3) an Nvidia Jetson TX2 for performing \ac{ANN} inference. The robot carries a depth sensor and an \ac{IMU} containing a 3-axis gyroscope and a 3-axis accelerometer. The primary sensing modality is vision using three global-shutter cameras. Two cameras are located in the front and are forward-facing, while a third camera is located in the back facing rearward with an externally mounted mirror used to effectively make the camera downward-facing. To improve the relative state estimation, we mounted an external magnetometer and a small downward-facing sonar used only to estimate the scale of our state estimator as described in Section \ref{sec:state_estimation}. The arrangement of the sensors is illustrated in Fig.~\ref{fig:robot_platform}.

Our interface to the robot is a short-range, two-way infrared remote control
that is used to start the experiments once underwater. A small screen on the robot is used to indicate which experiment will be run and provide feedback when configuring parameters. While experiments are being executed, ten multicolored LEDs on the top of the robot are used to indicate the status of the experiment and provide visual feedback to safety divers following the robot. The remote can be used to stop the robot and cancel the experiment if needed by the safety diver.

\subsection{State Estimation}
\label{sec:state_estimation}
State estimation is used in our experiments both for labelling data for training a goal-conditioned policy and for keeping track of the pose of the robot relative to a target waypoint. Our state estimation scheme is based on a combination of \acf{DSO} \cite{EngelDSo2017} and dead-reckoning for sections where visual features are insufficient for visual odometry.
The state estimation module processes images from the robot's downward-facing rear camera, as the robot's front cameras primarily view open water which contain few stable visual features. Conversely, the downward facing camera view is often dominated by coral which are rich in visual features.
Since \ac{DSO} does not resolve scale ambiguity itself, we need to estimate the scale on-line, as the scale ${}^{\mathbb{W}}_{\mathbb{O}}s_t$ between the odometry frame $\mathbb{O}$ and the world frame $\mathbb{W}$ drifts with time \cite{EngelDSo2017}.
To estimate ${}^{\mathbb{W}}_{\mathbb{O}}s_t$, we used the downward facing single beam sonar with aperture $\alpha = 30^{\circ}$.
We assume the sonar beam is projected from the optical center of the camera along the optical axis.
This ignores a translation of about \SI{4}{\cm} between the two.
Given the point cloud ${}^{\mathbb{C}}\mathcal{P}_t$ at time $t$ obtained by \ac{DSO} expressed in the camera frame $\mathbb{C}$, one can then select the points overlapping the latest sonar beam at time $t$:
\begin{equation}
{}^{\mathbb{C}}\mathcal{S}_t = \set*{\mathbf{p} \in {}^{\mathbb{C}}\mathcal{P}_t : \arccos\left({\frac{\mathbf{p}\cdot\mathbf{z}}{||\mathbf{p}||}}\right) \leq \alpha}
\end{equation}
where $\mathbf{z} = (0, 0, 1)$ is the direction of the optical axis in the camera frame.
Then, we compute the ratio between the average distance of those points to the optical center and the latest range reading of the sonar $r_t$:
\begin{equation}
{}^{\mathbb{W}}_{\mathbb{O}}\hat{s}_t = \frac{\sum_{\mathbf{p} \in {}^{\mathbb{C}}\mathcal{S}_t} ||\mathbf{p}||}{|{}^{\mathbb{C}}\mathcal{S}_t|r_t}
\end{equation}

This scale estimate could be corrupted by the noise in the sonar sensor and errors in the point cloud.
To reduce the variance, we use a time-smoothed estimate with horizon $H = 10$ computed as ${}^{\mathbb{W}}_{\mathbb{O}}\tilde{s}(t) = 1/H\sum_{\tau=t-H}^t {}^{\mathbb{W}}_{\mathbb{O}}\hat{s}_\tau$.
We can then scale the relative odometry pose ${}^{\mathbb{B}}_{\mathbb{O}}\mathcal{T}_{t-1}^{-1}{}^{\mathbb{B}}_{\mathbb{O}}\mathcal{T}_{t}$ of the robot frame $\mathbb{B}$ by ${}^{\mathbb{W}}_{\mathbb{O}}\tilde{s}(t)$ to get an estimate of the linear and angular velocity of the robot.

Finally, we fuse this velocity estimate with the \ac{IMU}, depth sensor and a constant forward speed prior in an \acf{EKF}. This constant bottom-speed prior, which is the speed at which the robot was always commanded to swim for our experiments, is used to compensate for the fact that \ac{DSO} may fail due to lack of visual features (for instance when swimming over sand). Since the robot was often subjected to substantial ocean currents, the water-speed generated by the robot's thrust
is not necessarily equal to its bottom-speed
relative to the sea floor.
Although the robot used a constant
thrust of roughly 0.41 m/s (0.8 knots), the surge and current had almost as large of an absolute magnitude for some sessions (which could be either forwards or backwards) so that the instantaneous bottom speed
varied substantially.

While \ac{DSO} is not running, the \ac{EKF} still fuses magnetometer, gyroscope, accelerometer, and depth sensor readings as well as a constant speed prior allowing for six degrees of freedom state estimation to function even in visually-deprived scenarios.

\subsection{Uncertainty Guided Exploration}
\label{sec:uncertainty}
While the behaviour policy provides data demonstrating the desired behaviour of the robot, it tends to not be very diverse. If the controller is working correctly, the robot will mostly swim in a straight line over coral while avoiding obstacles by pitching up. Furthermore, our real world datasets are relatively small: each deployment resulted in approximately 400 new samples for training the goal-conditioned policies. To make trajectories more diverse, while maintaining obstacle avoidance and coral following behaviours, we use an uncertainty weighting scheme based on the observation that robot actions tend to be more uncertain where action selection was not obvious to the labeller; e.g. in regions without obstacles. Thus, we can use the uncertainty of the action distribution to induce exploration safely. Since our model is reactive, if we merely sample from the action distribution the exploration behaviour would not be temporally consistent. To ensure temporal consistency, we periodically sample an exploration action and commit to it by a fixed duration. During this period, we use the uncertainty of the action predicted by the model for selecting between the exploration action and the predicted action.

More specifically, we sample an exploration action $\textbf{f}^{(\theta)}_{\textrm{expl}}$ from a categorical distribution with parameters $\big[p_{-3}^{\textrm{expl}}, p_{-2}^{\textrm{expl}}, p_{-1}^{\textrm{expl}}, p_{0}^{\textrm{expl}}, p_{1}^{\textrm{expl}}, p_{2}^{\textrm{expl}}, p_{3}^{\textrm{expl}}\big]$, and duration $T_{\textrm{expl}} \sim \mathrm{Uniform\left[T_{\textrm{lo}}, T_{\textrm{hi}}\right]}$. We evaluate the yaw action as predicted by the behaviour policy which corresponds to the parameters of a categorical distribution: $\textbf{f}^{(\theta)} = \left[p_{-3}, p_{-2}, p_{-1}, p_{0}, p_{1}, p_{2}, p_{3}\right]$. From this categorical distribution, we use its entropy  $\mathcal{H}(\textbf{f}^{(\theta)})$ as a measure of uncertainty. This entropy is used to compute a gating weight:
\begin{equation}
   w_{\mathcal{H}} = 1 - \mathrm{exp}\Big(-0.5\Big(\frac{\mathcal{H}(\textbf{f}^{(\theta)})}{B}\Big)^2 \Big)
\end{equation}
\noindent which is then used to compute the yaw action sent to the controller as
\begin{equation}
      \hat{\textbf{f}}^{(\theta)} = (1-w_{\mathcal{H}}) \textbf{f}^{(\theta)} + w_{\mathcal{H}} \textbf{f}^{(\theta)}_{\textrm{expl}}
\end{equation}
\noindent This sampling procedure is repeated when $T_{\textrm{expl}}$ seconds have passed, so that the robot commits to a different exploratory action.

\begin{figure}[t]
  \centering
  \includegraphics[width=1.0\linewidth]{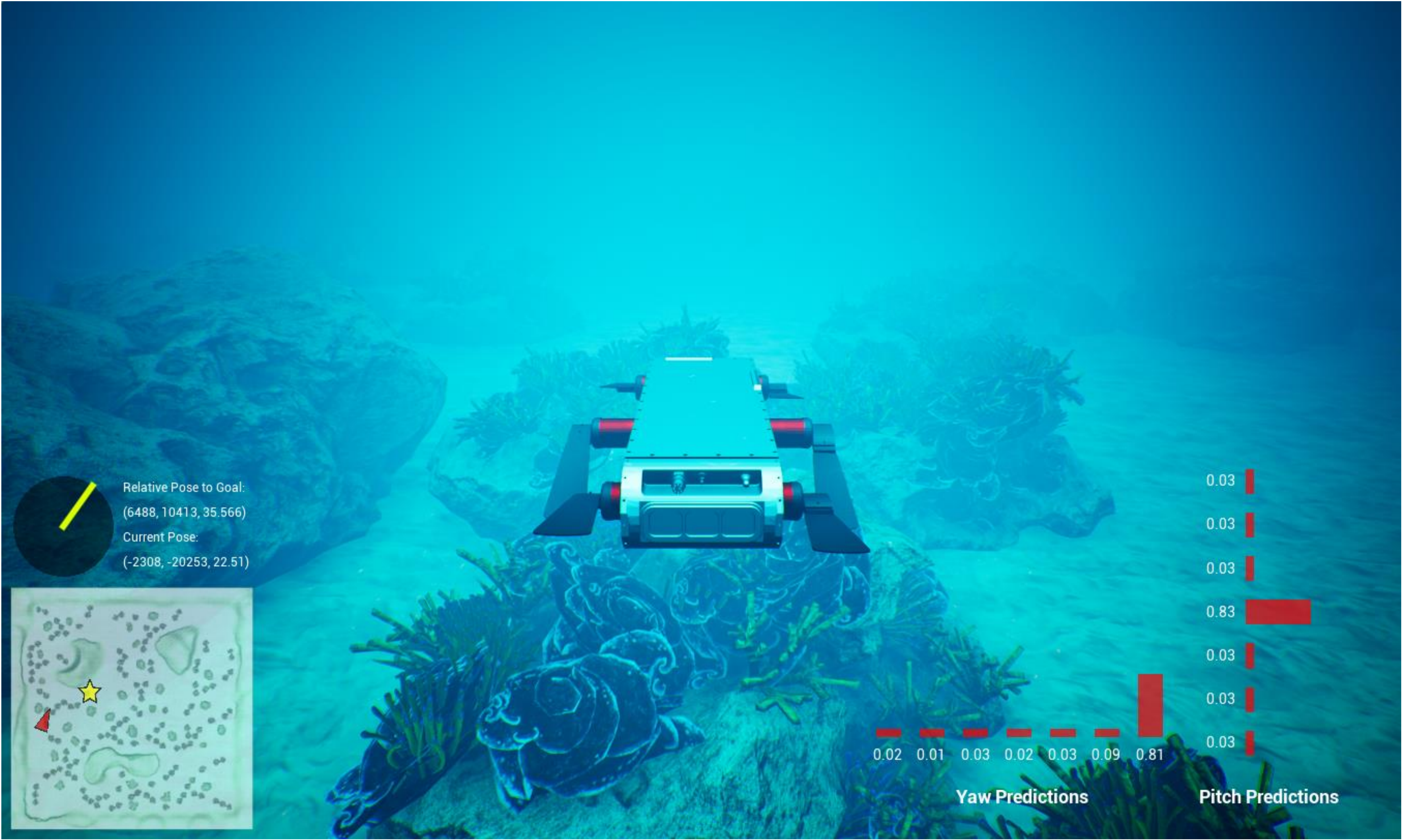}
  \caption{Screenshot of Unreal Engine \cite{unrealengine} simulation environment used for quantitative evaluation. }
\label{fig:simulator_overview}
\end{figure}
\begin{figure}[ht]
  \centering
  \includegraphics[width=1.0\linewidth]{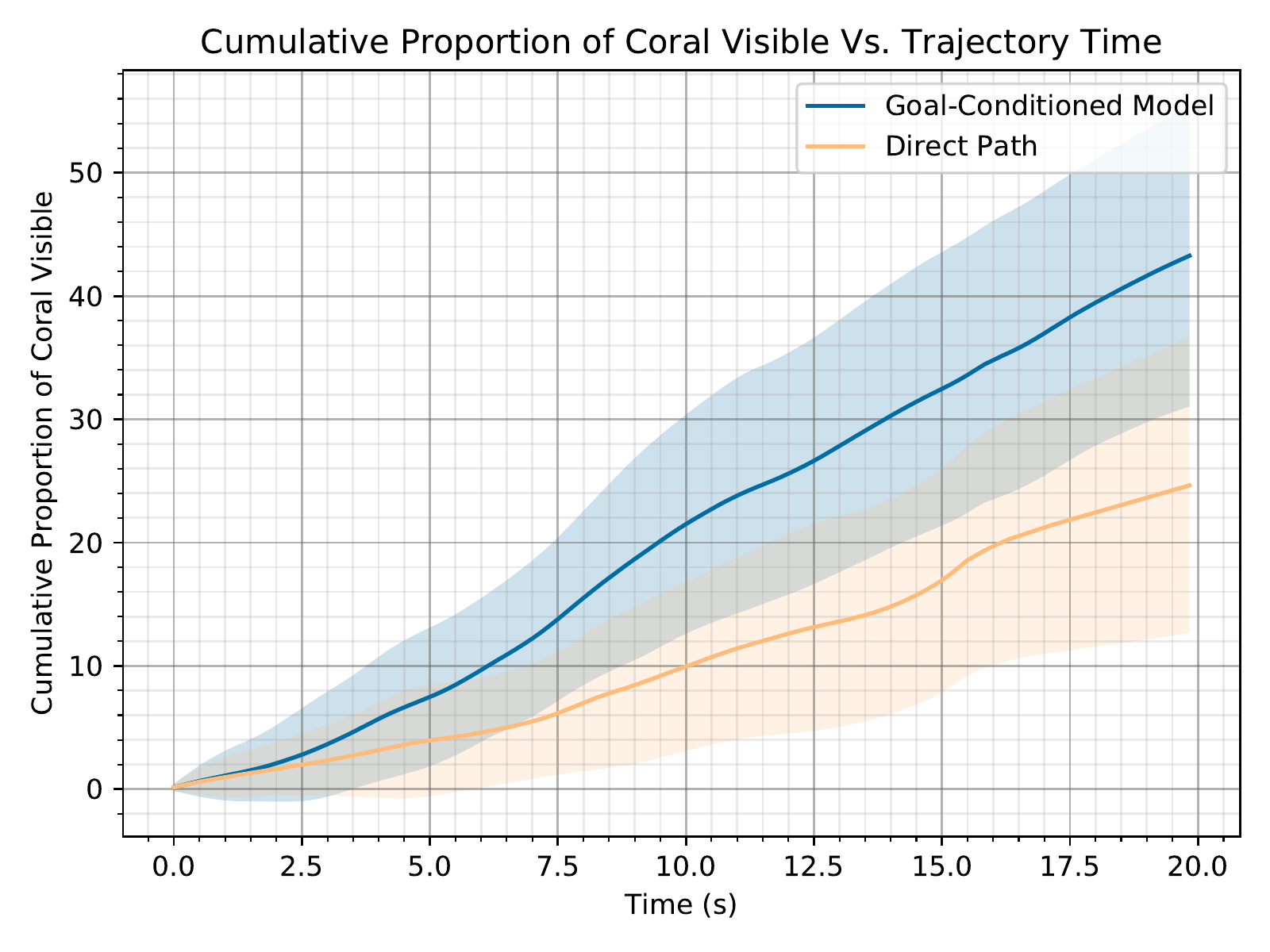}
  \vspace{-1em}
  \caption{Cumulative proportion of coral visible during trajectory execution. Results gathered from 30 randomized trials with varying start and goal positions. The robot swam between the two positions both directly and using our goal-conditioned policy. This demonstrates how goal-conditioned policy follows the underlying behaviour policy of staying over coral regions while simultaneously traveling towards a goal.}
\label{fig:coral_visible_trajectory}
\end{figure}

\begin{figure}[hb!]
  \centering
  \includegraphics[width=1.0\linewidth]{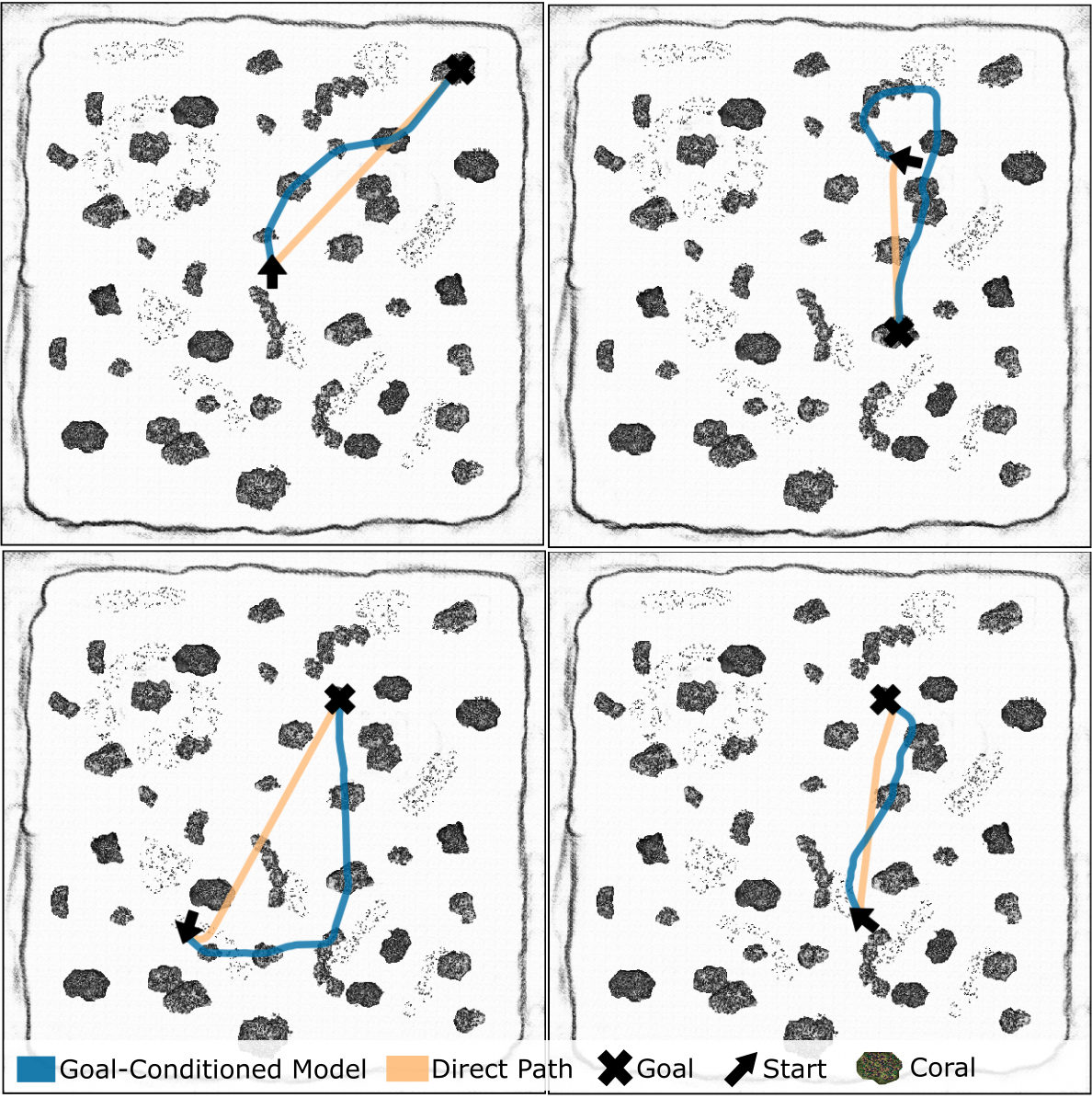}
  \caption{Sample trajectories from simulation illustrating the behavioural differences between the proposed goal conditioned model and a direct policy model which travels straight to the goal. The goal conditioned model maximizes coral visited along its path while simultaneously reaching the goal.}
\label{fig:simulation-sample-trajectories}
\end{figure}

\section{Experimental validation}
\label{sec:results}

Our experimental validation was conducted using two 
different approaches.  We used a simulation environment to obtain quantitative results in a repeatable, controlled, and consistent manner where performance with respect to ground-truth could be determined.  We used
deployment on an autonomous robot in the open ocean to
evaluate the real-world feasibility of our
approach, validate its utility and 
robustness, and generally show that it performs 
well in practice.  Each of these evaluation 
regimes provided a distinct and important kind
of validation.

\subsection{Simulation Results}
\label{sec:simulation}

Prior to real-world deployment, our model was validated in simulation. A custom underwater simulation environment with randomly placed coral was made using Unreal Engine \cite{unrealengine}. A screenshot of this environment is shown in Fig.~\ref{fig:simulator_overview}. To automate the data collection process for imitation learning training, an automated expert user was used to navigate to the nearest coral in the visible field of view by directly querying coral world positions on the map (information which is hidden to the learner). A simple PID controller was used to implement the expert's control policy.

Approximately 18,000 training samples composed of image, pose, and action tuples at 6 Hz control rate were gathered in simulation. Hindsight relabelling was applied to this dataset to generate goals for training as discussed in section~\ref{sec:hindsight_relabelling}.
After offline training, a model validation set accuracy of $79\%$ and $88\%$ was obtained for yaw and pitch respectively.

During online policy evaluation, performance or reward was measured as the amount of visible coral in the robot's field of view. Specifically, the percentage of visible coral at each timestep was calculated jointly from the forward-facing camera and downward-facing survey camera. Ray tracing was used to determine the proportion of interceptions on coral meshes in the current image frames and summed over all timesteps of a trajectory.

Fig.~\ref{fig:coral_visible_trajectory} compares the cumulative performance over the trajectory for the goal conditioned model versus a policy which takes the direct path to the goal. Fig.~\ref{fig:simulation-sample-trajectories} illustrates the difference in some sample trajectories. The results illustrate how the goal conditioned model simultaneously favours surveying coral while still navigating towards the goal in a reasonable amount of time.

\begin{figure}[t!]
  \centering
\includegraphics[width=1\linewidth]{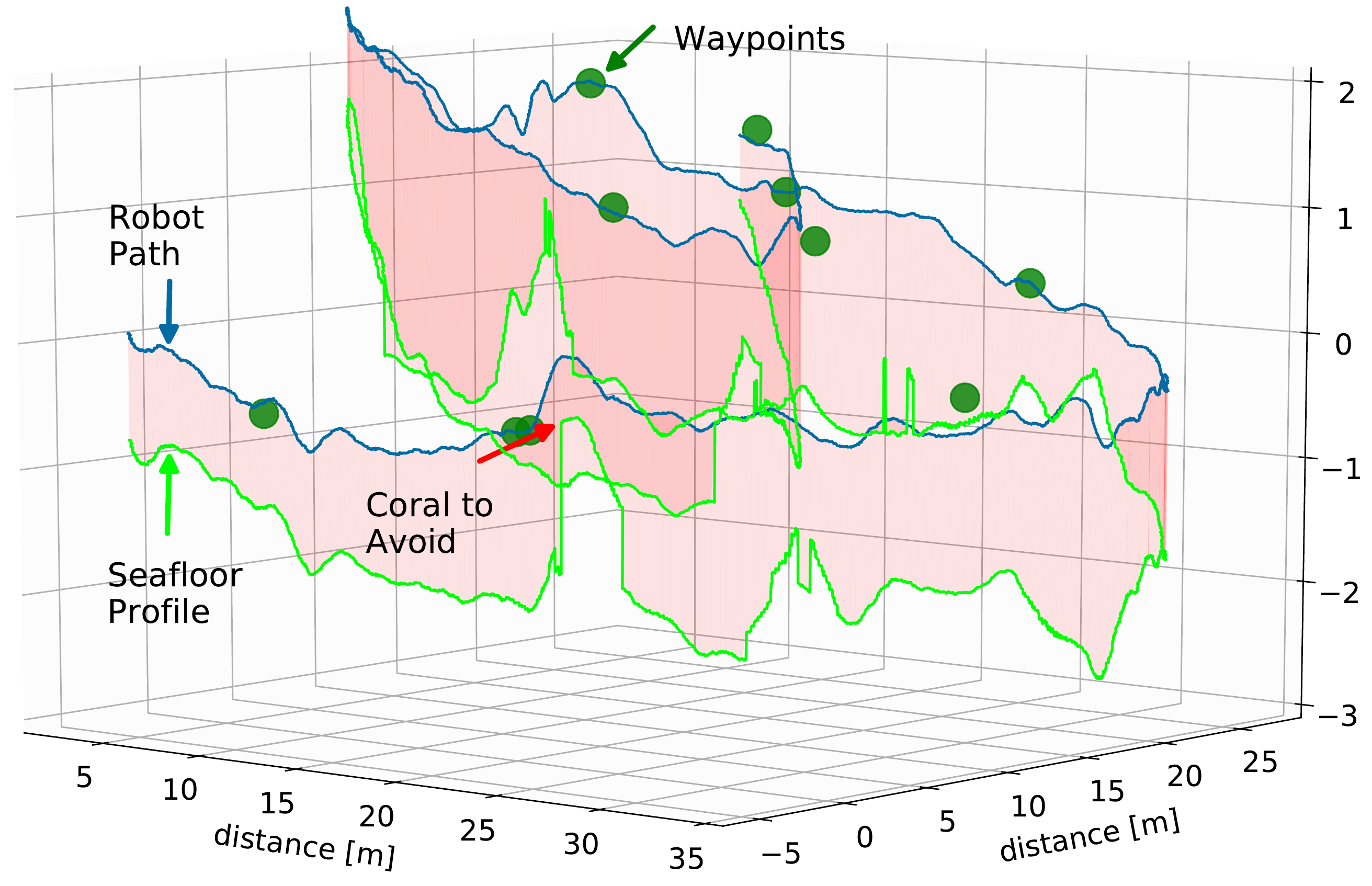}
  \caption{Example trajectory of the robot in the ocean navigating through several waypoints. Note that this is the same trajectory shown in Fig. \ref{fig:goal_conditioned_vs_originalbehaviour}. The robot travels towards the waypoints while also avoiding obstacles as reflected by the change in the robot path with respect to the seafloor profile.}
\label{fig:real_trajs}
\end{figure}

\begin{figure*}[ht!]
  \centering
  \includegraphics[width=1.0\linewidth]{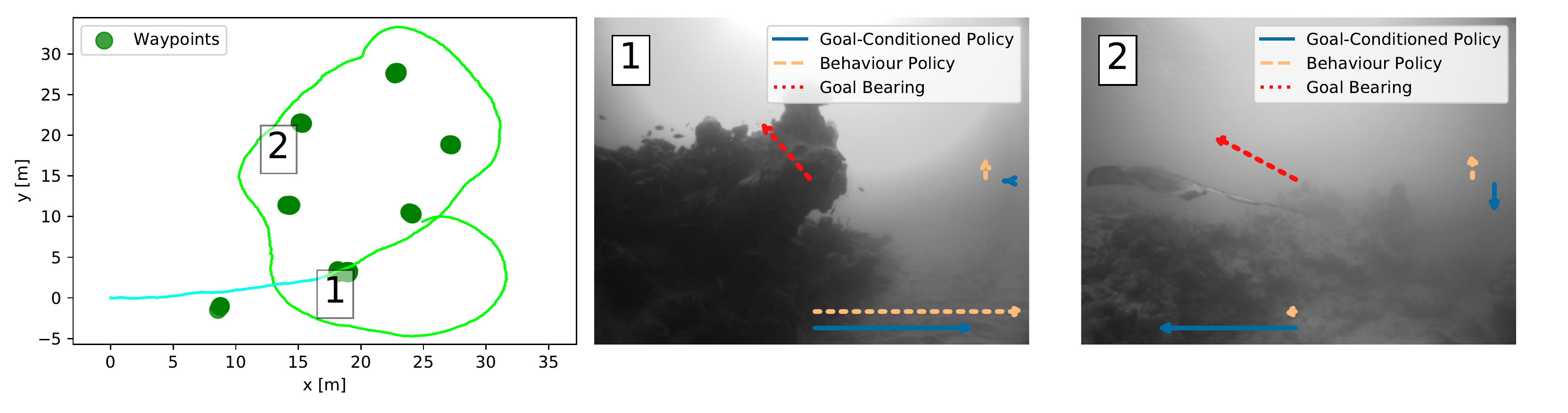} 
  \caption{Left: One of the trajectories executed by the robot, demonstrating the goal-reaching behaviour. Center: Obstacle avoidance overrides the goal-following behaviour (note the behaviour and goal-conditioned policies agree). Right: Goal-reaching overrides the original behaviour (both policies disagree) }
\label{fig:goal_conditioned_vs_originalbehaviour}
\end{figure*}

During evaluation, we compared Cartesian (x,y) and polar (magnitude, yaw) relative goal formats and found the Cartesian format to perform better. This may be due to the fact that periodic angle inputs might be difficult for the network to learn. Likewise, we experimented with both concatenation and multiplication when combining the goal with the ~\ac{CNN} output feature map as shown in Fig.~\ref{fig:gc_aqua_net}. It was found that the multiplication method yielded policies that were more sensitive to the goal position while those with concatenation had a tendency to sometimes ignore the goal which was undesirable.

\subsection{Open Ocean Deployment}
\label{sec:ocean}
We evaluated our system in the 
open ocean across a range of sea
conditions in the Caribbean Sea. The demanding character of operating in such an environment in the presence of surge, 
weather variations and lighting changes cannot be overstated.

We first validated a variant of the model of \cite{manderson2018vision} and collected new data for the behaviour policy. The dataset to train the behaviour policy consisted of 14,000 images from the initial deployments. To maintain consistency in the robot's behaviour, the whole dataset was relabelled by a single person. In this case, the labelling was done to try to keep the robot close to the coral reef while avoiding obstacles, with two differences to our related prior work~\cite{manderson2018vision}. First, the obstacle avoidance behaviour was biased to use pitch more often than yaw. Second, when there was no obvious yaw action (no obstacles or sand patches in front) the yaw label was selected with varying degrees of randomness\footnote{If only sand was visible, the yaw label was selected uniformly at random. If there was coral, but no obstacles, the action was selected to bring the coral towards the center of the images with varying degrees for the yaw action label}. This last difference allowed us to use the uncertainty-based exploration strategy described in Section~\ref{sec:uncertainty}. The re-trained model was deployed in the ocean, successfully replicating the results reported previously~\cite{manderson2018vision} and was used to obtain additional trajectories totalling approximately 20,500 samples for training the goal-conditioned navigation policy. In Fig.~\ref{fig:expl_traj} we illustrate a variety of trajectories we obtained during these deployments


During training of the goal-conditioned model, the collected dataset was split into training and validation sets using a 80/20 split. We found that multiplying the goal embedding vector with the feature vector from the \ac{CNN} provided the best convergence result: the validation accuracy was $81\%$ for yaw and $61\%$ for pitch. After training the model, we generated a set of waypoint trajectories spliced together using the method described in Section~\ref{sec:waypoint}.

Unlike the simulation results above, the real robot deployment presented a few challenges that made the task harder. First, the visibility varied greatly depending on the time of the day, the sky conditions, and the surge. Second, the surge made controlling the robot increasingly difficult, as the robot's motion is now influenced by 1) the relative position of the goal 2) the image captured by its front camera 3) external forces that were not explicitly modeled. Nevertheless, our method was able to guide the robot towards the goal. For the model that we are reporting the following results, we ran trials for 4 different 10-waypoint trajectories (with one illustrated in Fig.~\ref{fig:real_trajs}). For these trajectories, 1 was completed successfully in two different trials, 1 reached 8 out of 10 waypoints, while the other two reach 7 out of 10 waypoints. The trajectories that didn't finish were stopped early, using the infrared remote interface described in Section~\ref{sec:aqua}, due to surge conditions. As illustrated in Fig~\ref{fig:real_trajs} and the accompanying video, our method is able to produce a controller that successfully reaches the desired waypoints while being aligned with the behaviour policy.

The decisions taken by the goal-conditioned policy on these trajectories
are shown in Fig.~\ref{fig:goal_conditioned_vs_originalbehaviour}, along with the actions that would have been taken by the original behaviour policy without goal conditioning, as 
well as a greedy policy that takes an action proportional to the difference between the robot's yaw and the bearing to the next goal.

\section{Conclusion} 
\label{sec:conclusion}
We have demonstrated a robust method for visual navigation applied to underwater vehicles. Our method results in goal-conditioned controllers that allow for exploratory behaviour between waypoints, so that more scientifically-relevant observations may be recorded if available in the environment. This behavior specification is provided to the system in the form of labels for initial imitation learning of a behaviour policy. We have demonstrated the efficacy and robustness of our method both in simulation and in open ocean field trials, where the robot traversed nearly one kilometer in a coral reef ecosystem. 




\begin{acronym}
\acro{CIL}{Conditional Imitation Learning}
\acro{AUV}{Autonomous Underwater Vehicle}
\acro{DSO}{Direct Sparse Odometry}
\acro{EKF}{extended Kalman filter}
\acro{IMU}{Inertial Measurement Unit}
\acro{ANN}{Artificial Neural Network}
\acro{CNN}{Convolutional Neural Network}
\acro{SLAM}{simultaneous localization and mapping}
\acro{LED's}{Light-Emitting Diodes}
\end{acronym}

\bibliographystyle{plainnat}
\bibliography{references}

\end{document}